%% file: acl2020.tex
\documentclass[11pt,a4paper]{article}
\usepackage[hyperref]{acl2020}
\usepackage{times}
\usepackage{latexsym}
\usepackage{amsmath, amsthm, amssymb, amsfonts}
\usepackage{url}
\usepackage[disable]{todonotes}   

\newcommand{\GPTs}{\texttt{GPT-2S}}
\newcommand{\GPTm}{\texttt{GPT-2M}}
\newcommand{\GPTl}{\texttt{GPT-2L}}
\newcommand{\GPT}{\texttt{GPT-2}}

\aclfinalcopy 
\def\aclpaperid{2002} 

\title{GPT-too: A language-model-first approach for AMR-to-text generation}

\author{\parbox{\linewidth}{\centering
Manuel Mager{\rm\affmark[1]\thanks{~~This research was done during an internship at IBM Research AI.}}~~~~Ram\'on Fernandez Astudillo{\rm\affmark[2]}~~~~Tahira Naseem{\rm\affmark[2]}~~~~Md Arafat Sultan{\rm\affmark[2]}\\Young-Suk Lee{\rm\affmark[2]}~~~~Radu Florian{\rm\affmark[2]}~~~~Salim Roukos{\rm\affmark[2]}} \vspace{.12cm}
\\
\affaddr{\affmark[1] Institute for Natural Language Processing,\\ 
University of Stuttgart, Germany}\\
\affaddr{\affmark[2] IBM Research AI, Yorktown Heights, NY 10598, USA}\\
\affaddr{\texttt{manuel.mager@ims.uni-stuttgart.de}}\\
\affaddr{\texttt{\{ramon.astudillo,arafat.sultan\}@ibm.com}} \\
\affaddr{\texttt{\{tnaseem, ysuklee\}@us.ibm.com}}} 
\newcommand*{\affaddr}[1]{#1} 
\newcommand*{\affmark}[1][*]{\textsuperscript{#1}}

\date{}

\begin{document}
\maketitle
\begin{abstract}

Abstract Meaning Representations (AMRs) are broad-coverage sentence-level semantic graphs. Existing approaches to generating text from AMR have focused on training sequence-to-sequence or graph-to-sequence models on AMR annotated data only. In this paper, we propose an alternative approach that combines a strong pre-trained language model with cycle consistency-based re-scoring. Despite the simplicity of the approach, our experimental results show these models outperform all previous techniques on the English LDC2017T10 dataset, including the recent use of transformer architectures. In addition to the standard evaluation metrics, we provide human evaluation experiments that further substantiate the strength of our approach.

\end{abstract}

\section{Introduction}
\label{sec:intro}
\input{intro}

\section{Fine-tuning GPT-2 for conditional language generation}
\label{sec:Model}

In order to fine-tune a generative model (\GPT;~\newcite{radford2019language}) for conditional text generation, prior works fine-tune the language model to predict target text starting from the additional source text as context. In our experiments, we found it beneficial to fine-tune on the joint distribution of AMR and text instead i.e. also reconstruct the source. Given a tokenized sentence $w_1 \cdots w_N$ and the sequential AMR representation $a_1 \cdots a_M$ we maximized the joint probability

\begin{align*}
p_{\mbox{\scriptsize GPT-2}}(\mathbf{w}, \mathbf{a}) = &\prod_{j=1}^N p_{\mbox{\scriptsize GPT-2}}(w_j \mid w_{1:j-1}, a_{1:M})\\&\cdot \prod_{i=1}^M p_{\mbox{\scriptsize GPT-2}}(a_i \mid a_{1:i-1})   
\label{eq:reconstruction}
\end{align*}

A special separator token is added to mark the end of the sequential AMR representation. Special AMR symbols that should not be interpreted literally are assigned tokens from the \GPT~unused token list. In addition to this, we also observed that freezing the input embeddings when fine-tuning had positive impact in performance. 

At test time, we provide the AMR as context as in conventional conditional text generation:

\begin{equation}
\hat{w}_j = \arg\max_{w_j} \{p_{\mbox{\scriptsize GPT-2}}(w_j \mid w_{1:j-1}, a_{1:M}) \}\nonumber
\end{equation}

\section{Re-scoring via Cycle Consistency}

The general idea of cycle consistency is to assess the quality of a system's output based on how well an external `reverse' system can reconstruct the input from it. In previous works, cycle-consistency based losses have been used as part of the training objective in machine translation \cite{he2016dual} and speech recognition \cite{hori2019cycle}. It has also been used for filtering synthetic training data for question answering \cite{alberti-etal-2019-synthetic}. Here we propose the use of a cycle consistency measure to re-score the system outputs. 

In particular, we take the top $k$ sentences generated by our system from each gold AMR graph and parse them using an off-the-shelf parser to obtain a second AMR graph. We then re-score each sentence using the standard AMR parsing metric Smatch \cite{cai2013smatch} by comparing the gold and parsed AMRs.

\section{Experimental setup}
\label{sec:exp}

Following Previous works on AMR-to-text, we Use the standard LDC2017T10 AMR corpus for evaluation of the proposed model. This Corpus contains 36,521 training instances of AMR graphs in PENMAN notation and the corresponding texts. It also includes 1368 and 1371 development and test instances, respectively. We tokenize each input text using The JAMR toolkit \cite{flanigan2014discriminative}. The concatenation of an AMR graph and the corresponding text is split into words, special symbols and sub-word units using the \GPT~tokenizer. We add all arc labels seen in the training set and the root node \texttt{:root} to the vocabulary of the \GPT model, but we freeze the embedding layer for training. We use the Hugging Face implementation of \cite{wolf2019transformers} for \GPT~small (\GPTs), medium (\GPTm) and large (\GPTl). Fine-tuning converges after $6$ epochs, which takes just a few hours on a V100 GPU\footnote{Code for this paper is available at: \url{https://github.com/IBM/GPT-too-AMR2text}}. For cycle-consistency re-scoring we use an implementation of \newcite{naseem-etal-2019-rewarding} in PyTorch. For re-scoring experiments, we use a beam size of 15. 

\paragraph{AMR input representation.} we test three variants of AMR representation. First, a depth-first search (DFS) through the graph following \newcite{konstas2017neural}, where the input sequence is the path followed in the graph. Second, to see if GPT-2 is in fact learning from the graph structure, we remove all the edges from the DFS, keeping only the concept nodes. This has the effect of removing the relation information between concepts, such as subject/object relations. As a third option, we use the PENMAN representation without any modification. The three input representations are illustrated below:

\vspace{1mm}
\begin{tabular}{c p{5.5cm}}
\small Nodes   & \small \texttt{recommend advocate-01 it vigorous}\\
\small DFS     & \small \texttt{ recommend :ARG1 advocate-01 :ARG1 it :manner vigorous} \\
\small Penman     & \small \texttt{(r / recommend-01
  :ARG1 (a / advocate-01
          :ARG1 (i / it)
          :manner (v / vigorous)))}
\end{tabular}
\vspace{1mm}

\paragraph{Decoding.} 
For generation, we experiment with greedy decoding, beam search, and nucleus sampling \cite{holtzman2019curious}. For beam search, we explore beam sizes of $5$, $10$ and $15$. As the system, in some cases, produces repetitive output at the end of the text, we additionally perform a post-processing step to remove these occurrences. 

\paragraph{Metrics.} We considered the three automatic evaluation metrics commonly used in previous works. We compute BLEU \cite{papineni2002bleu} using  SacreBLEU \cite{ma2019results}. We compute chrF++ \cite{popovic2017chrf++} using both SacreBLEU and the scripts used by authors of the baseline systems. We compute METEOR \cite{banerjee2005meteor} with the default values for English of the CMU implementation.\footnote{\url{https://www.cs.cmu.edu/~alavie/METEOR}}

In addition to the standard automatic metrics, we also carry out human evaluation experiments and use the semantic similarity metric BERTScore \cite{bert-score}. Both metrics arguably have less dependency on the surface symbols of the reference text used for evaluation. This is particularly relevant for the AMR-to-text task, since one single AMR graph corresponds to multiple sentences with the same semantic meaning. Conventional metrics for AMR-to-text are are strongly influenced by surface symbols and thus do not capture well the ability of the system to produce a diverse sentences with same underlying semantics.

Human evaluations are carried out by three professional annotators on $51$ randomly selected sentences from the $1371$ test sentences, on a 6 point scale, ranging from 0 to 5.
\begin{itemize}
\item {\small 0=Exceptionally poor (No useful information is conveyed at all.)}
\item {\small 1=Poor (Fundamental errors in grammar and vocabulary make it difficult to understand the meaning.)}
\item {\small 2=Not good enough (Errors in grammar, vocabulary and style make it difficult to understand the meaning.)}
\item {\small 3=Good enough (There are errors in the text, but I am reasonably confident that I understand the meaning.)}
\item {\small 4=Very good (There may be minor errors in the text, but I am very confident that I understand the meaning.)}
\item {\small 5=Excellent (The information is presented clearly and with appropriate grammar, vocabulary and style.)}
\end{itemize}
For each system, scores from all annotators are averaged to compute a single score. Inter-annotator agreement was $0.7$ when measured by Pearson correlation coefficient.


Our system produces de-tokenized cased output after BPE decoding, whereas previous systems produce traditional tokenized lower-cased output. Therefore, we lowercase and tokenize our system outputs to have fair comparisons with previous systems.

\begin{table}[t]
    \centering
    \fontsize{10pt}{12pt}\selectfont
    \setlength{\tabcolsep}{2.0pt}
    \begin{tabular}{c|c||r|r}
 Model   & Input                          &BLEU & chrF++ \\ \hline \hline
\GPTs~Rec.& Only nodes AMR 	             &9.45 	& 41.59 \\
\GPTs~Rec.& Lin. AMR w/o edges.  &11.35 & 43.25 \\
\GPTs~Rec.& Lin. AMR w/edges. 	 &20.14 & 53.12 \\
\GPTs~Rec.& Penman AMR 	                 &22.37 & 53.92 \\ \hline
\GPTm~Rec.& Lin. AMR w/edges.  &22.86 & 55.04 \\
\GPTm~Rec.& Penman AMR 	 &27.99 & 61.26
    \end{tabular}
    \caption{Results on the LDC2017T10 development set using GPT-2 S(mall) and M(edium) with Rec(onstruction) loss (see \S2) for different AMR representations (see \S4).}
    
    \label{tab:representation_results}
\end{table}{}

\begin{table}[t]
    \centering
    \fontsize{10pt}{12pt}\selectfont
    \setlength{\tabcolsep}{2.0pt}
    \begin{tabular}{c|c||r|r}
Approach  & Decoding & BLEU & chrF++ \\ \hline \hline  
\GPTm~Conditional         &Greedy   & 25.73 & 57.2 \\\hline
\GPTm~Rec.              &Greedy   & 30.41 & 61.36 \\
\GPTm~Rec.              &BEAM     & 31.8 	& 62.56 \\
\GPTm~Rec.              &BEAM 10  & \textbf{32.32} & 62.79 \\
\GPTm~Rec.              &Sampling & 28.75 & 61.19
    \end{tabular}
    \caption{Results on the LDC2017T10 development set. Rec(onstruction) uses the AMR reconstruction term (see \S\ref{sec:Model}) whereas Conditional does not.}
    \label{tab:decoding_res}
\end{table}{}

\subsection{Results}
\label{subsec:results}

\begin{table}[h]
    \centering
    \fontsize{10pt}{12pt}\selectfont
    \setlength{\tabcolsep}{2.0pt}
    \begin{tabular}{c| c c c }
        System &  \multicolumn{3}{c}{Performance}\\\hline
               &  BLEU & Meteor & chrF++ \\\hline \hline
        \newcite{beck2018graph}         & 23.30     & -     & 50.40 \\
        \newcite{damonte2019structural} & 24.54     & 24.07 & - \\
        \newcite{guo2019densely}        & 27.60     & -     & 57.30 \\
        \newcite{cao2019factorising}    & 26.80     & -     & -  \\
        \newcite{sinh2019study}         & 18.36     & -     & - \\
        \newcite{ribeiro2019enhancing}  & 27.87     & 33.21 & - \\
        \newcite{Deng2020graph}         & 29.80     & 35.10 & 59.4  \\
        \newcite{zhu2019modeling}       & 31.82     & 36.38 & \textbf{64.05} \\
        \GPTm~Rec.                 & $32.10^\blacklozenge$      & $35.86^\Diamond$ & $61.81^\blacklozenge$ \\
        \GPTl~Rec.                 & $32.47^\blacklozenge$      & $36.80^\Diamond$  & $62.88^\blacklozenge$\\
        \GPTm~Rec. re-scoring       & $32.98^\blacklozenge$     & $37.33^\Diamond$ & $63.09^\blacklozenge$ \\
        \GPTl~Rec. re-scoring       & \bf 33.02$^\blacklozenge$ & \bf 37.68$^\Diamond$ & \bf 63.89$^\Box$ \\
    \end{tabular}
    \caption{Results on the LDC2017T10 test set for best performing models compared to other results reported in the literature. $^\blacklozenge$ indicates statistical significance at $(P < .01)$, $^\Diamond$ at $(P < 0.05)$ and $^\Box$, not significant. All significance tests are with respect to \citep{zhu2019modeling}.}
    \label{tab:compared_results}
\end{table}

Regarding the type of AMR representation, as shown in Table \ref{tab:representation_results}, using directly the PENMAN notation for AMR representation leads to the best results outperforming DFS. Edge information, indicating relations between concepts, seems also to play a fundamental role since its absence strongly decreases performance in both DFS and PENMAN representations. Penman notation was chosen for the rest of the experiments. 

The impact of the use of a reconstruction term explained in \S2 is shown in Table \ref{tab:decoding_res}. The model trained using this additional term achieves $30.41$ BLEU and $61.36$ chrF++, as opposed to $25.73$ BLEU and $57.2$ chrF++ without the term. We therefore use a reconstruction term training in the rest of the experiments.

Beam search improves system performance greatly over the greedy baseline with $1.91$ BLEU points (see Table \ref{tab:decoding_res}). With beam size $10$, we obtain $32.32$ BLEU and $62.79$ chrF++. With nucleus sampling at a cumulative probability mass of $0.9$, performance drops to $28.75$ BLEU and $61.19$ chrF++. Finally, cycle-consistency re-ranking of the beam search outputs improves performance ($33.57$ BLEU, $64.86$ chrF++) over the one best output. 

\begin{table}[]
    \centering
    \fontsize{10pt}{12pt}\selectfont
    \setlength{\tabcolsep}{3.0pt}
    \begin{tabular}{c|c c |c }
        System & \multicolumn{3}{c}{LDC2017T10}\\\hline
               & \multicolumn{2}{c|}{Human Eval.} & SemSim \\\hline 
               & Avg. & P45 & F1 \\\hline\hline
               
        \newcite{guo2019densely}       & $2.48$ & 15.69\% & 92.68  \\
        \newcite{ribeiro2019enhancing} & $2.42$ & 16.37\% & 92.63  \\
        \newcite{zhu2019modeling}      & $2.61$ & 20.26\% & 93.31  \\\hline
        \GPTm~Rec.                   &  $3.03$ & 37.91\% & 94.55 \\
        \GPTl~Rec.                   & \bf 3.04 & \bf 41.83\% & \bf 94.63  \\
    \end{tabular}
    \caption{Human evaluation and semantic similarity (SemSim) results on the LDC2017T10 test set. Human evaluations (Human Eval.) show the average (Avg.) of scores (0 to 5) and the ratio of sentence evaluated between 4 and 5 (P45). All results for human evaluation are on $51$ randomly selected sentences and statistically significant at $(P < 0.05)$. SemSim results are significant at $(P < 0.01)$. All significance tests refer to a comparison with \cite{zhu2019modeling}.}
    \label{tab:huaman_eval}
\end{table}

\begin{table*}[ht!]
    \centering
    \fontsize{10pt}{12pt}\selectfont
    \setlength{\tabcolsep}{2.5pt}
    \begin{tabular}{c r p{13cm}}

         \bf  & \bf System    & \bf Generated text\\\hline \hline
         (1) & \bf REF:    & the doctors gave her medication and it 's made her much better .\\
             & \bf G2S:    & the doctor gives her medications and they make her much better .\\ 
             & \bf Transf: & doctors give her medications and make her much better .\\
             & \bf Our:     & the doctor gave her the medication and made her feel much better. \\
             & \bf Our R.:& the doctor gave her the medication and made her " much better " .\\ \hline

(2) & \bf REF: & at the state scientific center of applied microbiology there is every kind of deadly bacteria that was studied for use in the secret biological weapons program of the soviet union . \\
& \bf G2S: & there are every kind of killing \textless unk\textgreater~in the state scientific center of applied microbiology to use themselves for soviet union 's secret biological weapons programs .  \\
& \bf Transf:  & there is every kind of bacterium , which is studied in using bacterium for the soviet union secret biological weapons program . \\
& \bf Our: & every kind of bacterium that was studied was found at the state scientific center of applied microbiology and was used in soviet secret weapons programs for biological weapons of biology . \\
& \bf Our R.: & every kind of bacterium that has been studied and used in soviet secret programs for biological weapons has been in the state scientific center of applied microbiology . \\\hline

         (3) & \bf REF:    & among the nations that have not signed the treaty only india and israel would qualify for admission to the nsg under the israeli proposal .\\
             & \bf G2S: & only one of the nations who do not sign the treaty are qualified for their proposal to admit the nsg .\\
             & \bf Transf:    & india and israel are only qualified for the nations that do not sign the treaty , but they admitted to the nsg .\\
             & \bf Our:    & india and israel are the only countries eligible to admit to the nsg by proposing a treaty .\\
             & \bf Our R.: & only india and israel are eligible to admit to the nsg by proposing a treaty .\\\hline
    \end{tabular}
    \caption{Output examples from four systems of the LDC2017T10 dataset. REF stands for reference, G2S for \citep{guo2019densely} and Transf. for \citep{zhu2019modeling}. Our is the top beam output for \GPTl~and Our R. is with re-scoring.}
    \label{tab:output_examples}
\end{table*}{}

Table \ref{tab:compared_results} compares the best \GPTm~and \GPTl~results, fine-tuned using the reconstruction term and PENMAN notation. For all scores we test statistical significance with a standard two-tailed student t-test. Our model achieves a large improvement of $1.2$ BLEU and $1.3$ METEOR scores over the previous state-of-the-art model using \GPTl~and re-scoring. For chrF++, we get different scores from SacreBLEU and the scripts provided by the authors of our baseline systems, achieving comparable results with the former ($63.89$), and improving over the best score with the latter ($65.01$) $(P < .01)$. 

Table \ref{tab:huaman_eval} shows human Evaluation results and semantic similarity scores of \GPTl~and \GPTm~compared to \cite{zhu2019modeling,ribeiro2019enhancing,guo2019densely}. Our approach produces a large number of high-quality sentences with $41.8\%$, a significant gain over the previous best system ($20.26\%$). Regarding semantic similarity, prior art methods show relatively close scores, a $0.9$ points difference, while \GPTl~Rec. improves $1.6$ points over the best of these models. It should be noted that differences with \cite{zhu2019modeling} for \GPTl~Rec. are statistically significantly with $P < .05$, while differences for \GPTm~Rec are not significant due to the small sample size. 

In Table \ref{tab:output_examples} we show three nontrivial examples, where we compare our system outputs with those of previous work. In the first example, the reference sentence contains a grammatical error. Our system not only generates the correct output, but also corrects the error in the reference. The proposed system can generate fluent long sentences as shown in example 2. The third example shows a sentence where all systems including ours fail to generate a correct text. 

\subsection{Discussion}
\label{subsec:discussion}

Due to the large amounts of data they are trained on, pre-trained transformer language models can be expected to generate fluent and diverse text \cite{see2019massively}. It should however be highlighted that fine-tuned \GPT~learns to produce not only fluent but also adequate text, despite using a sequential representation of an AMR graph as input. As shown in the experimental setup, encoding of relations plays as well a fundamental role in AMR-to-text performance, indicating that \GPT~attains a fine-grained understanding of the underlying semantics to reach state of the art performance.

While a sequence of PENMAN notation tokens is far from an optimal encoding of a graph, it is noteworthy how far performance-wise current strong language models can go. Furthermore, It is likely that standard metrics (BLEU, Meteor, chrF++) that rely on a reference text do not properly reflect AMR-to-text quality. An AMR graph corresponds to multiple sentences with the same semantics and these measures are likely biased towards the single available reference. In metrics that are less influenced by the reference text such as human evaluation and semantic similarity, the proposed system shows a larger improvement over the previous systems with close to $50\%$ of the generated sentences considered excellent or good. 

Finally it is worth considering that leveraging pre-trained transformers greatly expands the vocabulary available on AMR-to-text systems. A single AMR graph can correspond to multiple sentences with markedly different surface realizations, but manual annotation of AMR is a time consuming task. Approaches like the one proposed may be a simple solution for generation of diverse text data for AMR parser training or other applications were diversity play a role.

\section{Conclusions}

In this work, we present a language model-based approach for the AMR-to-text generation task. We show that a strong pre-trained transformer language model (\GPT) can be fine-tuned to generate text directly from the PENMAN notation of an AMR graph. Comparison with state-of-the-art models in BLUE, chrF++, METEOR as well as SemSim and human evaluation metrics show that while simple, this approach can outperform existing methods including methods training transformers from scratch. We also show that cycle consistency-based re-scoring using a conventional AMR parser and the Smatch metric can notably improve the results. Future work will focus on incorporating better encoding of the AMR graph into the current system and exploring data augmentation techniques leveraging the proposed approach.

\section*{Acknowledgments}

We thank the reviewers for their valuable suggestions. We would also like to thank Chunchuan Lyu for his valuable feedback and help.

\bibliography{acl2020}
\bibliographystyle{acl_natbib}

\end{document}

%% file: intro.tex
Abstract Meaning Representation (AMR) \cite{banarescu2013abstract} is a rooted, directed, acyclic graph with labeled edges (relations) and nodes (concepts) expressing ``who is doing what to whom''. AMR-to-text generates sentences representing the semantics underlying an AMR graph. 

Initial works in AMR-to-text used transducers ~\cite{flanigan2016generation}, phrase-based machine translation \cite{pourdamghani2016generating} and neural sequence-to-sequence (\texttt{seq2seq}) models with linearized graphs \cite{konstas2017neural}.
\newcite{cao2019factorising} leverage constituency parsing for generation. \newcite{beck2018graph} improve upon prior RNN graph encoding \cite{song2018graph} with Levi Graph Transformations. \newcite{damonte2019structural} compare multiple representations and find graph encoders to be the best. \newcite{guo2019densely} use RNN graph encoders with dense graph convolutional encoding. \newcite{ribeiro2019enhancing} use RNN encoders with dual graph representations.  Transformer-based \texttt{seq2seq} \cite{vaswani2017attention} was first applied to AMR-to-text in \cite{sinh2019study}. \newcite{zhu2019modeling} greatly improve over the prior state-of-the-art by modifying self-attention to account for AMR graph structure. Using transformers has also been recently explored by \newcite{wang2020amr} who propose a mutli-head graph attention mechanism and by \newcite{Deng2020graph} who propose a graph transformer architecture.

Pre-trained transformer representations \cite{radford2018improving,devlin2019bert,radford2019language} use transfer learning to yield powerful language models that considerably outperform the prior art. They have also shown great success when fine-tuned to particular text generation tasks \cite{see2019massively,zhang2019dialogpt,keskar2019ctrl}. Given their success, it would be desirable to apply pre-trained transformer models to a graph-to-text task like AMR-to-text, but the need for graph encoding precludes in principle that option. Feeding the network with some sequential representation of the graph, such as a topological sorting, looses some of the graphs representational power. Complex graph annotations, such as AMR, also contain many special symbols and special constructs that departure from natural language and may by not interpretable by a pre-trained language model.

In this paper we explore the possibility of directly fine-tuning a pre-trained transformer language model on a sequential representation of AMR graphs, despite the expected difficulties listed above. For this we re-purpose a GPT-2 language model \cite{radford2019language} to yield an AMR-to-text system. We show that it is surprisingly easy to fine-tune GPT-2 to learn AMR graph to text mapping that outperforms the previous state-of-the-art on automatic evaluation metrics. Since a single graph AMR, graph corresponds to multiple sentences with the same meaning, we also provide human evaluation and semantic similarity metric results \cite{bert-score} which are less dependent on reference text. Human evaluation and semantic similarity results highlight the positive impact of a strong language model strategy. Finally we also introduce a simple re-scoring technique based on cycle-consistency that further improves performance.

%% file: acl2020.bbl
\begin{thebibliography}{34}
\expandafter\ifx\csname natexlab\endcsname\relax\def\natexlab#1{#1}\fi

\bibitem[{Alberti et~al.(2019)Alberti, Andor, Pitler, Devlin, and
  Collins}]{alberti-etal-2019-synthetic}
Chris Alberti, Daniel Andor, Emily Pitler, Jacob Devlin, and Michael Collins.
  2019.
\newblock Synthetic {QA} corpora generation with roundtrip consistency.
\newblock In \emph{Proceedings of the 57th Annual Meeting of the Association
  for Computational Linguistics}.

\bibitem[{Banarescu et~al.(2013)Banarescu, Bonial, Cai, Georgescu, Griffitt,
  Hermjakob, Knight, Koehn, Palmer, and Schneider}]{banarescu2013abstract}
Laura Banarescu, Claire Bonial, Shu Cai, Madalina Georgescu, Kira Griffitt, Ulf
  Hermjakob, Kevin Knight, Philipp Koehn, Martha Palmer, and Nathan Schneider.
  2013.
\newblock Abstract meaning representation for sembanking.
\newblock In \emph{Proceedings of the 7th Linguistic Annotation Workshop and
  Interoperability with Discourse}, pages 178--186.

\bibitem[{Banerjee and Lavie(2005)}]{banerjee2005meteor}
Satanjeev Banerjee and Alon Lavie. 2005.
\newblock Meteor: An automatic metric for mt evaluation with improved
  correlation with human judgments.
\newblock In \emph{Proceedings of the acl workshop on intrinsic and extrinsic
  evaluation measures for machine translation and/or summarization}, pages
  65--72.

\bibitem[{Beck et~al.(2018)Beck, Haffari, and Cohn}]{beck2018graph}
Daniel Beck, Gholamreza Haffari, and Trevor Cohn. 2018.
\newblock Graph-to-sequence learning using gated graph neural networks.
\newblock In \emph{Proceedings of the 56th Annual Meeting of the Association
  for Computational Linguistics (Volume 1: Long Papers)}, pages 273--283.

\bibitem[{Cai and Lam(2020)}]{Deng2020graph}
Deng Cai and Wai Lam. 2020.
\newblock Graph transformer for graph-to-sequence learning.
\newblock In \emph{34th AAAI conference on artificial intelligence}.

\bibitem[{Cai and Knight(2013)}]{cai2013smatch}
Shu Cai and Kevin Knight. 2013.
\newblock Smatch: an evaluation metric for semantic feature structures.
\newblock In \emph{Proceedings of the 51st Annual Meeting of the Association
  for Computational Linguistics (Volume 2: Short Papers)}, pages 748--752.

\bibitem[{Cao and Clark(2019)}]{cao2019factorising}
Kris Cao and Stephen Clark. 2019.
\newblock Factorising amr generation through syntax.
\newblock In \emph{Proceedings of the 2019 Conference of the North American
  Chapter of the Association for Computational Linguistics: Human Language
  Technologies, Volume 1 (Long and Short Papers)}, pages 2157--2163.

\bibitem[{Damonte and Cohen(2019)}]{damonte2019structural}
Marco Damonte and Shay~B Cohen. 2019.
\newblock Structural neural encoders for amr-to-text generation.
\newblock In \emph{Proceedings of the 2019 Conference of the North American
  Chapter of the Association for Computational Linguistics: Human Language
  Technologies, Volume 1 (Long and Short Papers)}, pages 3649--3658.

\bibitem[{Devlin et~al.(2019)Devlin, Chang, Lee, and
  Toutanova}]{devlin2019bert}
Jacob Devlin, Ming-Wei Chang, Kenton Lee, and Kristina Toutanova. 2019.
\newblock Bert: Pre-training of deep bidirectional transformers for language
  understanding.
\newblock In \emph{Proceedings of the 2019 Conference of the North American
  Chapter of the Association for Computational Linguistics: Human Language
  Technologies, Volume 1 (Long and Short Papers)}, pages 4171--4186.

\bibitem[{Flanigan et~al.(2016)Flanigan, Dyer, Smith, and
  Carbonell}]{flanigan2016generation}
Jeffrey Flanigan, Chris Dyer, Noah~A Smith, and Jaime Carbonell. 2016.
\newblock Generation from abstract meaning representation using tree
  transducers.
\newblock In \emph{Proceedings of the 2016 conference of the North American
  chapter of the association for computational linguistics: Human language
  technologies}, pages 731--739.

\bibitem[{Flanigan et~al.(2014)Flanigan, Thomson, Carbonell, Dyer, and
  Smith}]{flanigan2014discriminative}
Jeffrey Flanigan, Sam Thomson, Jaime Carbonell, Chris Dyer, and Noah~A Smith.
  2014.
\newblock A discriminative graph-based parser for the abstract meaning
  representation.
\newblock In \emph{Proceedings of the 52nd Annual Meeting of the Association
  for Computational Linguistics (Volume 1: Long Papers)}, pages 1426--1436.

\bibitem[{Guo et~al.(2019)Guo, Zhang, Teng, and Lu}]{guo2019densely}
Zhijiang Guo, Yan Zhang, Zhiyang Teng, and Wei Lu. 2019.
\newblock Densely connected graph convolutional networks for graph-to-sequence
  learning.
\newblock \emph{Transactions of the Association for Computational Linguistics},
  7:297--312.

\bibitem[{He et~al.(2016)He, Xia, Qin, Wang, Yu, Liu, and Ma}]{he2016dual}
Di~He, Yingce Xia, Tao Qin, Liwei Wang, Nenghai Yu, Tie-Yan Liu, and Wei-Ying
  Ma. 2016.
\newblock Dual learning for machine translation.
\newblock In \emph{Advances in Neural Information Processing Systems}, pages
  820--828.

\bibitem[{Holtzman et~al.(2019)Holtzman, Buys, Forbes, and
  Choi}]{holtzman2019curious}
Ari Holtzman, Jan Buys, Maxwell Forbes, and Yejin Choi. 2019.
\newblock The curious case of neural text degeneration.
\newblock \emph{arXiv preprint arXiv:1904.09751}.

\bibitem[{Hori et~al.(2019)Hori, Astudillo, Hayashi, Zhang, Watanabe, and
  Le~Roux}]{hori2019cycle}
Takaaki Hori, Ramon Astudillo, Tomoki Hayashi, Yu~Zhang, Shinji Watanabe, and
  Jonathan Le~Roux. 2019.
\newblock Cycle-consistency training for end-to-end speech recognition.
\newblock In \emph{ICASSP 2019-2019 IEEE International Conference on Acoustics,
  Speech and Signal Processing (ICASSP)}, pages 6271--6275. IEEE.

\bibitem[{Keskar et~al.(2019)Keskar, McCann, Varshney, Xiong, and
  Socher}]{keskar2019ctrl}
Nitish~Shirish Keskar, Bryan McCann, Lav~R Varshney, Caiming Xiong, and Richard
  Socher. 2019.
\newblock Ctrl: A conditional transformer language model for controllable
  generation.
\newblock \emph{arXiv preprint arXiv:1909.05858}.

\bibitem[{Konstas et~al.(2017)Konstas, Iyer, Yatskar, Choi, and
  Zettlemoyer}]{konstas2017neural}
Ioannis Konstas, Srinivasan Iyer, Mark Yatskar, Yejin Choi, and Luke
  Zettlemoyer. 2017.
\newblock Neural amr: Sequence-to-sequence models for parsing and generation.
\newblock In \emph{Proceedings of the 55th Annual Meeting of the Association
  for Computational Linguistics (Volume 1: Long Papers)}, pages 146--157.

\bibitem[{Ma et~al.(2019)Ma, Wei, Bojar, and Graham}]{ma2019results}
Qingsong Ma, Johnny Wei, Ond{\v{r}}ej Bojar, and Yvette Graham. 2019.
\newblock Results of the wmt19 metrics shared task: Segment-level and strong mt
  systems pose big challenges.
\newblock In \emph{Proceedings of the Fourth Conference on Machine Translation
  (Volume 2: Shared Task Papers, Day 1)}, pages 62--90.

\bibitem[{Naseem et~al.(2019)Naseem, Shah, Wan, Florian, Roukos, and
  Ballesteros}]{naseem-etal-2019-rewarding}
Tahira Naseem, Abhishek Shah, Hui Wan, Radu Florian, Salim Roukos, and Miguel
  Ballesteros. 2019.
\newblock Rewarding {S}match: Transition-based {AMR} parsing with reinforcement
  learning.

\bibitem[{Papineni et~al.(2002)Papineni, Roukos, Ward, and
  Zhu}]{papineni2002bleu}
Kishore Papineni, Salim Roukos, Todd Ward, and Wei-Jing Zhu. 2002.
\newblock Bleu: a method for automatic evaluation of machine translation.
\newblock In \emph{Proceedings of the 40th Annual Meeting of the Association
  for Computational Linguistics}, pages 311--318.

\bibitem[{Popovi{\'c}(2017)}]{popovic2017chrf++}
Maja Popovi{\'c}. 2017.
\newblock chrf++: words helping character n-grams.
\newblock In \emph{Proceedings of the second conference on machine
  translation}, pages 612--618.

\bibitem[{Pourdamghani et~al.(2016)Pourdamghani, Knight, and
  Hermjakob}]{pourdamghani2016generating}
Nima Pourdamghani, Kevin Knight, and Ulf Hermjakob. 2016.
\newblock Generating english from abstract meaning representations.
\newblock In \emph{Proceedings of the 9th international natural language
  generation conference}, pages 21--25.

\bibitem[{Radford et~al.(2018)Radford, Narasimhan, Salimans, and
  Sutskever}]{radford2018improving}
Alec Radford, Karthik Narasimhan, Tim Salimans, and Ilya Sutskever. 2018.
\newblock Improving language understanding by generative pre-training. 2018.
\newblock \emph{URL https://s3-us-west-2. amazonaws.
  com/openai-assets/research-covers/language-unsupervised/language\_understanding\_paper.
  pdf}.

\bibitem[{Radford et~al.(2019)Radford, Wu, Child, Luan, Amodei, and
  Sutskever}]{radford2019language}
Alec Radford, Jeffrey Wu, Rewon Child, David Luan, Dario Amodei, and Ilya
  Sutskever. 2019.
\newblock Language models are unsupervised multitask learners.
\newblock \emph{OpenAI Blog}, 1(8).

\bibitem[{Ribeiro et~al.(2019)Ribeiro, Gardent, and
  Gurevych}]{ribeiro2019enhancing}
Leonardo~FR Ribeiro, Claire Gardent, and Iryna Gurevych. 2019.
\newblock Enhancing amr-to-text generation with dual graph representations.
\newblock \emph{arXiv preprint arXiv:1909.00352}.

\bibitem[{See et~al.(2019)See, Pappu, Saxena, Yerukola, and
  Manning}]{see2019massively}
Abigail See, Aneesh Pappu, Rohun Saxena, Akhila Yerukola, and Christopher~D
  Manning. 2019.
\newblock Do massively pretrained language models make better storytellers?
\newblock \emph{arXiv preprint arXiv:1909.10705}.

\bibitem[{Sinh and Le~Minh(2019)}]{sinh2019study}
Vu~Trong Sinh and Nguyen Le~Minh. 2019.
\newblock A study on self-attention mechanism for amr-to-text generation.
\newblock In \emph{International Conference on Applications of Natural Language
  to Information Systems}, pages 321--328. Springer.

\bibitem[{Song et~al.(2018)Song, Zhang, Wang, and Gildea}]{song2018graph}
Linfeng Song, Yue Zhang, Zhiguo Wang, and Daniel Gildea. 2018.
\newblock A graph-to-sequence model for amr-to-text generation.
\newblock In \emph{Proceedings of the 56th Annual Meeting of the Association
  for Computational Linguistics (Volume 1: Long Papers)}, pages 1616--1626.

\bibitem[{Vaswani et~al.(2017)Vaswani, Shazeer, Parmar, Uszkoreit, Jones,
  Gomez, Kaiser, and Polosukhin}]{vaswani2017attention}
Ashish Vaswani, Noam Shazeer, Niki Parmar, Jakob Uszkoreit, Llion Jones,
  Aidan~N Gomez, {\L}ukasz Kaiser, and Illia Polosukhin. 2017.
\newblock Attention is all you need.
\newblock In \emph{Advances in neural information processing systems}, pages
  5998--6008.

\bibitem[{Wang et~al.(2020)Wang, Wan, and Jin}]{wang2020amr}
Tianming Wang, Xiaojun Wan, and Hanqi Jin. 2020.
\newblock Amr-to-text generation with graph transformer.
\newblock \emph{Transactions of the Association for Computational Linguistics},
  8:19--33.

\bibitem[{Wolf et~al.(2019)Wolf, Debut, Sanh, Chaumond, Delangue, Moi, Cistac,
  Rault, Louf, Funtowicz et~al.}]{wolf2019transformers}
Thomas Wolf, Lysandre Debut, Victor Sanh, Julien Chaumond, Clement Delangue,
  Anthony Moi, Pierric Cistac, Tim Rault, R{\'e}mi Louf, Morgan Funtowicz,
  et~al. 2019.
\newblock Transformers: State-of-the-art natural language processing.
\newblock \emph{arXiv preprint arXiv:1910.03771}.

\bibitem[{Zhang et~al.(2020)Zhang, Kishore, Wu, Weinberger, and
  Artzi}]{bert-score}
Tianyi Zhang, Varsha Kishore, Felix Wu, Kilian~Q. Weinberger, and Yoav Artzi.
  2020.
\newblock \href {https://openreview.net/forum?id=SkeHuCVFDr} {Bertscore:
  Evaluating text generation with bert}.
\newblock In \emph{International Conference on Learning Representations}.

\bibitem[{Zhang et~al.(2019)Zhang, Sun, Galley, Chen, Brockett, Gao, Gao, Liu,
  and Dolan}]{zhang2019dialogpt}
Yizhe Zhang, Siqi Sun, Michel Galley, Yen-Chun Chen, Chris Brockett, Xiang Gao,
  Jianfeng Gao, Jingjing Liu, and Bill Dolan. 2019.
\newblock Dialogpt: Large-scale generative pre-training for conversational
  response generation.
\newblock \emph{arXiv preprint arXiv:1911.00536}.

\bibitem[{Zhu et~al.(2019)Zhu, Li, Zhu, Qian, Zhang, and
  Zhou}]{zhu2019modeling}
Jie Zhu, Junhui Li, Muhua Zhu, Longhua Qian, Min Zhang, and Guodong Zhou. 2019.
\newblock Modeling graph structure in transformer for better amr-to-text
  generation.
\newblock In \emph{Proceedings of the 2019 Conference on Empirical Methods in
  Natural Language Processing and the 9th International Joint Conference on
  Natural Language Processing (EMNLP-IJCNLP)}, pages 5462--5471.

\end{thebibliography}
